\documentclass[journal]{IEEEtran}
\usepackage{amsmath,amsfonts}
\usepackage{algorithm}
\usepackage{algpseudocode}
\usepackage{array}
\usepackage{stfloats}

\usepackage{xcolor}
\usepackage{url}
\usepackage{verbatim}
\usepackage{graphicx}
\usepackage{subfig} % Load the subfig package
\usepackage{svg}
\usepackage{cite}
\usepackage{booktabs} % for horizontal rules
\usepackage{multirow} % for multirow cells
\usepackage{array} % for better column formatting
\usepackage{tabularx} % for auto-width columns
\usepackage{algpseudocode} % Pseudocode styling
\usepackage{listings} % For source code listing
\begin{document}

\title{ Hyperspectral Images Efficient Spatial and Spectral non-Linear Model with Bidirectional Feature Learning}

\author{{Judy X~Yang,~\IEEEmembership{Student Member,~IEEE},
Jing~Wang,
Zekun~Long,~\IEEEmembership{Student Member,~IEEE},
Chenhong Sui, ~\IEEEmembership{Senior Member,~IEEE}
Jun~Zhou,~\IEEEmembership{Senior Member,~IEEE}}  % <-this % stops a space
\thanks{Judy X Yang, Jun Zhou, and Zekun Long are with the School of Information and Communication Technology, Griffith University, Australia (corresponding author: Jun Zhou, jun.zhou@griffith.edu.au).}% <-this % stops a space
\thanks{Jing Wang is with the Department of Agriculture and Fisheries, Queensland Government, Australia.}
\thanks{Chenhong Sui is with the School of Information Science and Engineering, Yantai University, China.}
}

\maketitle

\begin{abstract}
Classifying hyperspectral images (HSIs) is a complex task in remote sensing due to the high-dimensional nature and volume of data involved. To address these challenges, we propose the Spectral-Spatial non-Linear Model, a novel framework that significantly reduces data volume while enhancing classification accuracy. Our model employs a bidirectional reversed convolutional neural network (CNN) to efficiently extract spectral features, complemented by a specialized block for spatial feature analysis. This hybrid approach leverages the operational efficiency of CNNs and incorporates dynamic feature extraction inspired by attention mechanisms, optimizing performance without the high computational demands typically associated with transformer-based models.
The SS non-Linear Model is designed to process hyperspectral data bidirectionally, achieving notable classification and efficiency improvements by fusing spectral and spatial features effectively. This approach yields superior classification accuracy compared to existing benchmarks while maintaining computational efficiency, making it suitable for resource-constrained environments. We validate the SS non-Linear Model on three widely recognized datasets, Houston 2013, Indian Pines, and Pavia University, demonstrating its ability to outperform current state-of-the-art models in HSI classification and efficiency.
This work highlights the innovative methodology of the SS non-Linear Model and its practical benefits for remote sensing applications, where both data efficiency and classification accuracy are critical. For further details, please refer to our code repository on GitHub: HSILinearModel. 
\end{abstract}


 % Main text
\section{Introduction}\label{Introduction}

\sloppy
Hyperspectral imaging (HSI) has emerged as a transformation tool in remote sensing, enabling highly detailed land cover analysis and environmental monitoring. Unlike traditional imaging methods, HSI captures hundreds of contiguous narrow spectral bands throughout the electromagnetic spectrum~\cite{wang2023hyperspectral, ran2016bands}. This capability allows for the precise detection and classification of surface materials by capturing subtle spectral characteristics that are often indistinguishable from conventional RGB images~\cite{ikeuchi2021computer, mateen2018role}. The granular spectral information offered by HSI has propelled its use in a wide array of applications, including precision agriculture, urban planning, mineral exploration, and environmental monitoring~\cite{barbedo2023review, tao2017automatic, vadrevu2022remote}.

However, the high-dimensionality inherent in hyperspectral data poses considerable challenges for classification tasks. The vast number of spectral bands increases computational complexity, requiring effective methods that can reduce data volume while preserving critical spectral and spatial information~\cite{moharram2023land}. An approach to mitigating these challenges is band selection~\cite{cai2019bs,wang2019attend,yang2024lidar}, a technique that involves selecting a subset of informative spectral bands. Band selection aims to retain only the most discriminative spectral features, which helps reduce dimensionality and computational burden without compromising classification accuracy. This approach has proven valuable in applications where computational efficiency is critical, such as real-time processing or analysis on resource-constrained platforms~\cite{sun2019hyperspectral}.

In parallel, attention-based models such as Vision Transformers (ViTs)\cite{vaswani2017attention, aleissaee2023transformers} have made significant strides in visual representation learning, excelling in embedding global context into each segment of an image. ViTs have shown marked improvements over traditional CNNs in various domains\cite{hong2021spectralformer, he2019hsi}. However, the memory-intensive nature and high computational requirements of ViTs remain substantial limitations, particularly in applications constrained by hardware and power considerations, such as remote sensing.

Recent developments in state-space models (SSMs) present a potential solution to these limitations. By enabling parallel processing and efficiently capturing long-range dependencies, SSMs offer a promising alternative to transformer-based models in computationally intensive tasks. The Mamba model exemplifies this approach, showcasing linear scalability and competitive performance in vision applications~\cite{gu2023mamba, zhu2024vision}. Drawing inspiration from both the computational efficiency of the Mamba model and the successful feature extraction capabilities of CNNs, we envision a model that blends these strengths to address the unique complexities of HSI data.

In response, we introduce the Spectral-Spatial non-linear (SS non-linear) model, a novel architecture designed to enhance HSI classification through a fusion of CNN and attention-based strategies. Our model employs a bidirectional network to efficiently process hyperspectral data, integrating both spectral and spatial information into a unified representation. By leveraging CNNs for spatial feature extraction alongside Mamba-inspired mechanisms for reduced computational load, the SS non-linear Model achieves robust classification performance with lower memory and computational costs. This architecture supports a comprehensive understanding of both local and global patterns in HSI data, effectively addressing the high-dimensional nature of hyperspectral images without the intensive computational overhead associated with transformers.

The SS non-linear Model was rigorously evaluated on three prominent hyperspectral 
datasets—Houston 2013, Indian Pines, and Pavia University—demonstrating superior classification performance and computational efficiency compared to current state-of-the-art models.  By reducing GPU memory usage, CPU load, and inference time, the SS non-linear Model represents a promising advancement in HSI analysis, particularly for dense prediction tasks that benefit from efficient processing pipelines.

Our contributions are summarized as follows:
\begin{itemize} 
\item We introduce the SS non-linear Model, a novel bidirectional framework for HSI classification that integrates spatial and spectral feature extraction. This model uniquely balances computational efficiency and classification accuracy, setting a new standard for HSI classification architectures. 
\item The SS non-linear Model achieves computational advantages over existing models, such as, RNN, CNN-based architectures, and ViT-based. This design minimizes memory and processing requirements, facilitating scalable classification suitable for practical remote sensing applications.
\item We validate the performance of the SS non-linear model through extensive experiments on three major datasets: Houston 2013, Indian Pines, and Pavia University. Our results consistently outperform vision transformer benchmarks, establishing the SS non-linear Model as a high-performance, efficient option for hyperspectral image classification. \end{itemize}

The rest of this paper is structured as follows. Section 2 reviews related work, situating our approach within the current landscape of HSI classification research. Section 3 details the methodology of the SS non-linear model, while Section 4 presents experimental results and analyzes. Finally, Section 5 concludes with insight into potential future research directions.

\section{Related Work}\label{sec:Related Work}
\slash
This section reviews recent advances in hyperspectral image (HSI) classification, focusing on deep learning models that leverage CNNs, transformers, and state-space models (SSMs). Each category brings unique strengths to the task of HSI classification, contributing to the development of increasingly efficient and accurate models.
\subsection{Deep Learning in Hyperspectral Image Classification}\label{sec:DLCNN}
The application of deep learning to HSI classification has brought substantial improvements in the extraction and analysis of spectral and spatial features. Various architectures, including CNNs, recurrent neural networks (RNNs), and generative adaptive networks (GANs), have been customized to the unique characteristics of HSI data~\cite{zhou2019learning, yu2017convolutional, mou2017deep, zhu2018generative, paoletti2018capsule}. Among these, CNNs have been especially impactful, demonstrating their ability to capture localized spatial features essential for precise classification.

CNN-Based Models: CNNs have been central to HSI classification due to their ability to process data in a hierarchical structure, capturing complex spatial and spectral information. Early CNN approaches for HSI classification, such as 2-D CNN~\cite{yang2018hyperspectral}, utilized multiple 2D convolutional layers to analyze spatial features, coupled with pooling and batch normalization layers to control dimensionality and improve generalization. Building on this, the R-2D-CNN model introduced residual connections, enhancing gradient flow and allowing deeper network architectures that maintain spatial resolution across layers.

The 3-D CNN model~\cite{yang2018hyperspectral} expanded on this approach by integrating 3D convolutional blocks, which enable the model to capture both spectral and spatial features simultaneously. This capability allows for a more nuanced understanding of hyperspectral data, where spectral correlations across adjacent bands are essential for accurate classification. However, while 3-D CNNs capture spectral dependencies, their computational demands are relatively high due to the volumetric nature of 3D convolutions, which can limit their scalability.

Advanced CNN Models: M3D-DCNN~\cite{he2017multi} was introduced to overcome some of the limitations of early CNN models by jointly learning 2D multi-scale spatial features and 1D spectral features in a unified framework. This end-to-end approach allows M3D-DCNN to better handle the high-dimensional nature of hyperspectral data while reducing computational load, making it particularly effective on large-scale HSI datasets. Multi-scale techniques in M3D-DCNN enable the network to adapt to various feature resolutions, capturing both fine-grained and broader contextual information within hyperspectral images.

While CNNs have proven effective for spatial feature extraction, they have some limitations in fully capturing spectral dependencies across a large number of bands. Models that focus solely on CNN architectures may overlook long-range spectral relationships, which are essential in hyperspectral analysis. This gap has led to the integration of CNNs with other architectures, such as transformers and RNNs, to improve feature representation across both spatial and spectral dimensions.

RNN-Based Models for Spectral Dependencies: RNNs, such as those developed by Mou et al.~\cite{mou2017deep}, address some of the limitations of CNNs by focusing on sequential data processing, making them well suited for capturing spectral dependencies across bands. Mou et al. introduced the parametric rectified tanh (PRetanh) activation function alongside modified gated recurrent units to handle spectral sequences in HSIs effectively. However, RNNs have their own limitations as their sequential processing can be computationally intensive, especially for large-scale datasets.

\subsection{Transformers  in Hyperspectral Image Classification}\label{sec: Transformer}
The advent of transformers has marked a significant development in HSI classification, leveraging self-attention mechanisms to model long-range dependencies within spectral data. Originally developed for NLP, Vision Transformers (ViTs)~\cite{vaswani2017attention} have been adapted for vision tasks, including hyperspectral classification, where the capture of global context in spectral bands is crucial.

Regarding transformer-based models, models such as SpectralFormer~\cite{hong2021spectralformer} have utilized transformer architectures to capture spectral information through cross-layer skip connections, effectively enhancing feature extraction without requiring conventional preprocessing steps. HSI-BERT~\cite{he2019hsi} introduced bidirectional transformers to model spectral-spatial dependencies, leading to improvements in classification accuracy and generalization.

More recent transformer models, such as Deep ViT~\cite{zhou2021deepvit}, T2T~\cite{yuan2021tokens}, LeViT~\cite{graham2021levit}, and HiT~\cite{yang2022hyperspectral}, have further advanced HSI classification by integrating spectral-spatial modeling capabilities. These models demonstrate the versatility of transformers in handling HSI data, particularly by capturing both local and global dependencies through self-attention.

Hybrid Transformer-CNN Models: The integration of CNNs with transformers has also led to the development of hybrid architectures that harness the strengths of both approaches. For example, the hyperspectral image transformer (HiT)\cite{yang2022hyperspectral} employs a CNN-based convolution block within the transformer framework to capture local spatial information, using spectral adaptive 3D convolutions for improved feature representation. Other hybrid models, such as the multiscale convolutional transformer by Jia et al.\cite{jia2022multiscale} and the spectral-spatial feature tokenization transformer (SSFTT) by Sun et al.~\cite{sun2022spectral}, demonstrate how CNN-transformer fusion can enhance the representation of spectral-spatial features, improving classification results in HSI tasks.

Although transformers have achieved significant performance gains, their high computational demands present challenges for practical deployment, especially in resource-constrained environments. This limitation has motivated the search for more efficient alternatives that can maintain accuracy without excessive resource consumption.

\subsection{The Emergence of State Space Models: Mamba in Computer Vision Applications}
State Space Models (SSMs) have recently gained traction in vision applications as a scalable alternative to transformers. The Mamba model~\cite{gu2023mamba}, a prominent example of an SSM-based approach, has demonstrated non-linear scalability and effective handling of long-range dependencies, making it suitable for tasks that require efficient processing across large data volumes. By replacing attention mechanisms with a computationally efficient backbone, Mamba offers a promising solution to the high resource demands of transformer models.

Mamba’s linear computational scalability has shown particular promise in high-resolution imagery and tasks that involve complex dependency modelling, such as HSI classification~\cite{fu2022hungry}. The incorporation of SSMs into CNN-based architectures further enhances model efficiency, offering a viable path to address the resource-intensive nature of deep learning in HSI tasks. Mamba’s success in various benchmarks, with reduced memory and computational demands, suggests a new direction for HSI classification frameworks.

The integration of CNNs, transformers, and SSMs represents a significant evolution in the field of HSI classification. CNNs have provided a strong foundation for capturing local spatial features and dimensionality reduction, with models like 3-D CNN and M3D-DCNN pushing the boundaries of spatial-spectral feature extraction. Transformer-based approaches, such as SpectralFormer and HiT, have introduced a paradigm shift by capturing long-range dependencies and global context, although their computational demands limit widespread adoption. Emerging SSM-based models, such as Mamba, offer a scalable and efficient alternative, demonstrating strong potential for high-resolution and resource-efficient HSI classification.

Together, these advancements underscore the importance of combining different model architectures to balance computational efficiency with classification accuracy. As HSI classification continues to develop, the synergy between CNNs, transformers, and SSMs is likely to drive further innovations in remote sensing, enabling more sophisticated and accessible solutions to analyze complex hyperspectral data.

\section{Methodology}

This section presents the proposed SS non-linear Model methodology, designed specifically for hyperspectral image (HSI) classification. The SS non-linear Model aims to address two critical challenges in HSI analysis: (1) high computational demands due to the large volume of spectral and spatial data and (2) maintaining or improving classification accuracy. By leveraging a unique bidirectional processing framework, our model effectively reduces the computational resources required for HSI classification while simultaneously enhancing performance.

The SS non-linear Model is built upon core principles of spectral-spatial feature extraction, guided by non-linear transformation parameters \( A \) and \( B \), which enable efficient forward and backward processing along the spectral dimension. The model’s architecture integrates a bidirectional Bi-Networks block for spectral analysis and a spatial processing block for spatial context capture. Together, these components create a comprehensive representation of hyperspectral data, optimized for efficient processing and improved classification.

The following sections describe the mathematical foundations, architectural components, and computational benefits of the SS non-linear Model in detail.

\subsection{SS non-Linear Model Model  Preliminaries}
The SS non-linear Model is built upon bidirectional spectral processing principles, specifically tailored for hyperspectral image (HSI) classification. The model leverages two transformation matrices,\( A \) and \( B \), designed to capture forward and backward dependencies across the spectral dimension. This bidirectional approach enables the model to synthesize spectral information both sequentially and retrospectively, thus creating a richer feature representation. 

The central components of this model,\( A \) and \( B \), govern the dynamics of the forward and backward processing paths, respectively. For each spectral band \( t \), the model processes hyperspectral data \( x(t) \) as follow: 
\begin{itemize}
    \item {\textbf{Forward Spectral Dependency}}The transformation parameter 
\( A \in \mathbb{R}^{N \times N} \) 
  captures forward dependencies, enabling the model to process information in the natural order of spectral bands. This forward progression is beneficial for retaining spectral continuity and building a detailed representation of spectral features. 
    
    \item {\textbf{Backward Spectral Dependency}}  Conversely, the transformation parameter 
\( B \in \mathbb{R}^{N \times N} \)
  is responsible for capturing dependencies in the reverse direction, allowing the model to integrate information from later spectral bands back to earlier bands. This backward processing enriches the model by synthesizing spectral information that complements the forward path, ensuring that feature representation accounts for both prior and subsequent bands.
\end{itemize}

\subsubsection{Bidirectional Processing Framework}
In the SS non-linear Model, hyperspectral data  \( x(t) \) at each spectral band \( t \), is projected into a hidden state representation 
\( h(t) \) that integrates both forward and backward information. The forward and backward state transitions are modeled as follows:

\begin{equation}
h_{\text{forward}}(t) = f(A \cdot x(t) + h_{\text{forward}}(t-1))
\label{eq:forward_state}
\end{equation}

\begin{equation}
h_{\text{backward}}(t) = f(B \cdot x(t) + h_{\text{backward}}(t+1))
\label{eq:backward_state}
\end{equation}

where \( f \) is a non-linear activation function (e.g., SiLU or tanh), applied to the hidden states to enhance the feature representation.

This bidirectional processing not only captures sequential dependencies, but also improves the discriminative power of the extracted features. By integrating forward and backward information, the SS non-linear Model effectively encapsulates the spectral structure in hyperspectral data, which is critical for accurate classification.
\subsubsection{Discrete Transformations for Spectral Sequencing}
The parameters \( A \) and \( B \) further facilitate discrete transformations, specifically designed to handle the spectral sequencing inherent in hyperspectral images. These transformations aim to amplify the discriminative capacity of the spectral features, improving the model's performance in classifying hyperspectral data. By encoding both spectral directions, the model achieves a comprehensive feature set that surpasses traditional unidirectional approaches in representation power.

The discrete transformation operations can be mathematically described as follows:

\begin{equation}
    y_{\text{forward}}(t) = g(A \cdot x(t))
\end{equation}

\begin{equation}
    y_{\text{backward}}(t) = g(B \cdot x(t))
\end{equation}

where \( g\)  is an additional transformation function that serves to enhance the separability of features, further helping classification tasks. The final spectral feature representation 
$(y\_{\text{combined}}$ is obtained by averaging or summing the outputs from the forward and backward paths:

\begin{equation}
y_{\text{combined}} = \text{reduce}(y_{\text{forward}} + y_{\text{backward}})
\label{eq:combined_output1}
\end{equation}

This combination effectively captures a holistic view of spectral information, thus setting a robust foundation for spatial processing and classification stages.

\subsection{Proposed Method Overview and Architecture Description}

\begin{figure*}[!ht]
\centering
\includegraphics[width=17cm, height=7cm]{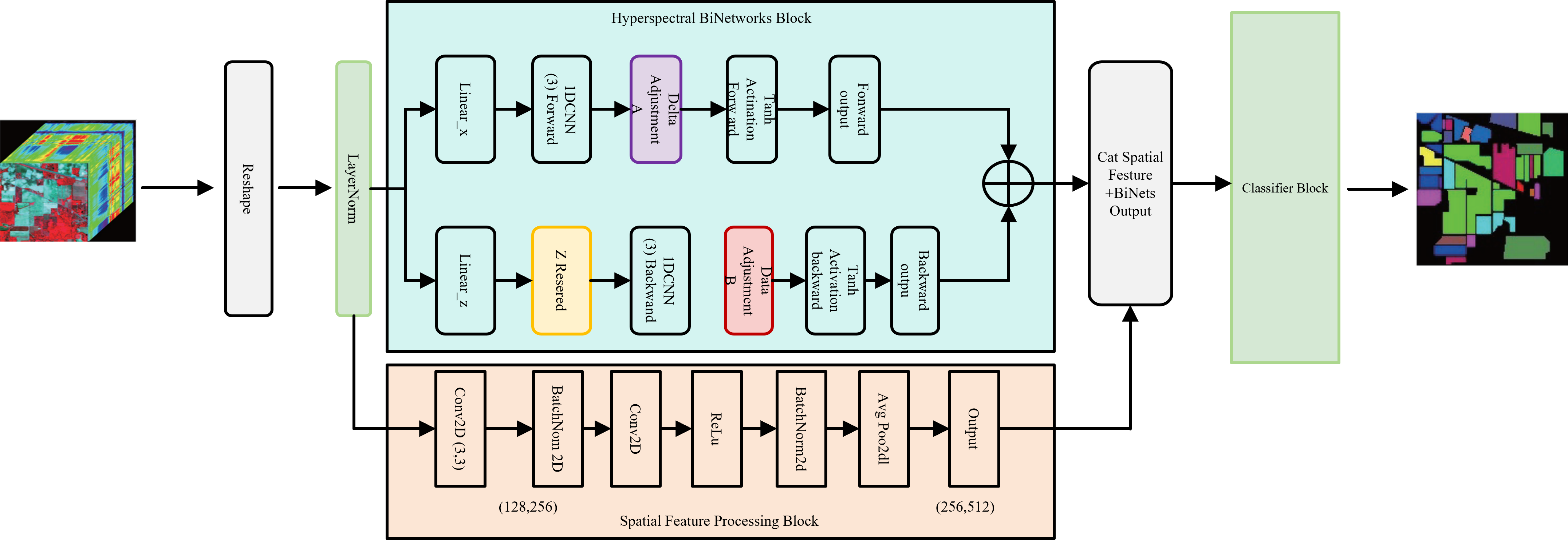} % Update the path to your image
\caption{The architectural overview of the Proposed SS non-linear Model model. The framework consists of four main components: (A) A hyperspectral image patch with dimensions \( p \times p \times \text{CH} \); (B) The Hyperspectral Bi-networks Block; (C) The Spatial Processing Block; (D) The classifier. The process begins by extracting patches that serve as input to the SS non-linear Model block. This block includes a spatial processing stage that precedes the unique forward and backward operations, and the concatenation between the Bi-networks output and the spatial feature process is the input of the classifier block. }
\label{hsivim}
\end{figure*}

The proposed SS non-linear Model architecture (illustrated in Figure~\ref{hsivim}) is designed to efficiently handle the high-dimensional data of hyperspectral images by dividing the processing into distinct blocks that target spectral and spatial information separately. This architecture consists of four main components:
1. Input Patch Preparation; 2. Hyperspectral Bi-Networks Block (for bidirectional spectral processing); 3. Spatial Feature Processing Block; 4. Classifier Block. 

The workflow begins with pre-processing the hyperspectral input data into patches, followed by spectral and spatial feature extraction. Finally, these features are combined and fed into a classifier to predict the class labels.

\subsubsection{Input Patch Preparation}
The hyperspectral input data \( x \)  is first divided into small patches of size 
\( p \times p \times \text{CH} \), where 
\( p \) is the spatial dimension and \( CH \) represents the number of spectral bands. Each patch is normalized using Layer Normalization to stabilize the inputs and improve convergence. The normalized data is then projected into higher-dimensional hidden states for further processing.
\subsubsection{Hyperspectral BiNetwork Block}

The hyperspectral bi-network block is central to the architecture of the SS non-linear Model. It processes the spectral information bidirectionally, utilizing forward and backward transformations to capture spectral dependencies in both directions. The following steps outline the processing within the Bi-Networks Block:

\textbf{Forward Projection:} The input data 
\( x \) is projected using a linear transformation to create 
\( x_{\text{proj}} \) , which serves as the initial hidden state for forward processing.

\textbf{Backward Projection:} The input data 
\( x \) is also projected into \( z_{\text{proj}} \) and reversed along the spectral dimension to form  \( z_{\text{proj\_reversed}} \),which initializes the backward processing path.

The forward and backward hidden states evolve through 1D convolutional layers and activation functions as follows:

\begin{equation}
x_{\text{forward}} = f(\text{Conv1d}(x_{\text{proj}}))
\label{eq:forward_conv}
\end{equation}

\begin{equation}
x_{\text{backward}} = f(\text{Conv1d}(z_{\text{proj\_reversed}}))
\label{eq:backward_conv}
\end{equation}

A delta-modulated non-linearity is applied to each pathway to enhance feature representation. The updated states in each direction are computed with transformation matrices \( A \) and \( B \):

\begin{equation}
h_{\text{forward}} = \tanh(x_{\text{forward}} + A \cdot \Delta_{\text{expanded}})
\label{eq:forward_state_update}
\end{equation}

\begin{equation}
h_{\text{backward}} = \tanh(x_{\text{backward}} + B \cdot \Delta_{\text{expanded}})
\label{eq:backward_state_update}
\end{equation}

The final output of the Bi-Networks Block, \( h_{\text{combined}} \), is obtained by reducing (e.g., averaging) the hidden states in each direction:

\begin{equation}
h_{\text{combined}} = \text{reduce}(h_{\text{forward}}) + \text{reduce}(h_{\text{backward}})
\label{eq:combined_output2}
\end{equation}

\subsubsection{Spatial Feature Processing Block}
The spatial structure within each patch undergoes further convolution and non-linearity in the Spatial Feature Processing Block \( h_{\text{spatial}} \). This block captures the spatial dependencies within the data, enhancing the feature representation obtained from the Bi-Networks Block. The resulting spatial feature representation is concatenated with 
\( h_{\text{combined}} \) to form a comprehensive feature vector.
\begin{equation}
h_{\text{spatial}} = f_{\text{spatial}}(\text{Conv2d}(h_{\text{combined}}))
\label{eq:spatial_feature}
\end{equation}

\begin{equation}
h_{\text{final}} = \text{concat}(h_{\text{spatial}}, h_{\text{combined}})
\label{eq:final_feature_vector}
\end{equation}

\subsubsection{Classifier Block}
The concatenated features from the Bi-Networks Block and the Spatial Feature Processing Block are passed to a classifier block, which performs the final classification. This block comprises a series of fully connected layers and activation functions, culminating in a softmax layer that produces the predicted class labels for each patch.

\begin{equation}
y_{\text{class}} = \text{softmax}(W \cdot h_{\text{final}} + b)
\label{eq:classifier_output}
\end{equation}

\subsubsection{Algorithm for SS non-linear Model Operation}
The SS non-linear Model efficiently processes hyperspectral image patches by integrating bidirectional spectral feature extraction and spatial processing to achieve accurate classification. The main steps in this operation involve Input Normalization, Forward and Backward Projections,Bidirectional Spectral Processing,Spatial Feature Processing, and Feature Concatenation and Classification.

\begin{algorithm}[!ht]
\caption{SS non-linear Model Operation}
\begin{algorithmic}[1]
\Require Hyperspectral patch \( x \in \mathbb{R}^{p \times p \times \text{CH}} \)
\Ensure Classification label \( \hat{y} \) for the input patch

\State \textbf{Input Normalization:} Normalize \( x \) using LayerNorm to obtain \( x_{\text{norm}} \)

\State \textbf{Forward and Backward Projections:} 
    \State Project \( x_{\text{norm}} \) to \( x_{\text{proj}} \) and \( z_{\text{proj}} \) using non-linear transformations
    \State Reverse \( z_{\text{proj}} \) along the spectral dimension to obtain \( z_{\text{proj\_reversed}} \)

\State \textbf{Bidirectional Spectral Processing:}
    \State Compute forward path: \( x_{\text{forward}} = f(\text{Conv1d}(x_{\text{proj}})) \)
    \State Compute backward path: \( x_{\text{backward}} = f(\text{Conv1d}(z_{\text{proj\_reversed}})) \)
    \State Apply delta modulation:
    \[
    h_{\text{forward}} = \tanh(x_{\text{forward}} + A \cdot \Delta_{\text{expanded}})
    \]
    \[
    h_{\text{backward}} = \tanh(x_{\text{backward}} + B \cdot \Delta_{\text{expanded}})
    \]
    \State Combine hidden states: \( h_{\text{combined}} = \text{reduce}(h_{\text{forward}}) + \text{reduce}(h_{\text{backward}}) \)

\State \textbf{Spatial Feature Processing:}
    \State Extract spatial features using convolutional layers to obtain \( h_{\text{spatial}} \)
    \State Concatenate \( h_{\text{combined}} \) and \( h_{\text{spatial}} \)

\State \textbf{Classification:}
    \State Pass concatenated features to classifier block
    \State Output classification label \( \hat{y} \)

\end{algorithmic}
\end{algorithm}

\subsection{Architectural Efficiency Analysis}
The SS non-linear Model is designed to reduce computational demands while maintaining or improving classification accuracy. This section provides a quantitative analysis of the model’s efficiency, highlighting its advantages in terms of FLOPs, runtime, and overall computational complexity.

\subsubsection{Floating Point Operations (FLOPs) Reduction}
FLOPs provide a measure of the computational load associated with performing a single forward pass in a model. In hyperspectral image classification, conventional CNNs and transformers often involve high FLOPs due to extensive convolutional or self-attention operations across multiple spectral bands and spatial dimensions.
The SS non-linear Model achieves a substantial reduction in FLOPs by adopting a bidirectional spectral processing approach without extensive kernel operations. Instead of using large convolutional filters or complex attention mechanisms, the SS non-linear Model leverages simpler 1D convolutions along with the efficient transformation matrices \( A \) and \( B \)  resulting in lower computational complexity.
To summarize, for a hyperspectral image with dimensions \( Batch \times H \times W \times C \), where:
\begin{itemize}
    \item \( Batch \) is the batch size,
    \item \( H \) and \( W \) are spatial dimensions,
    \item \( CH \) is the spectral dimension (number of bands),
\end{itemize}

The FLOPs for a forward pass in different models can be approximated as follows:

\begin{itemize}
    \item Transformer Model: 
      \begin{itemize}
        \item Parameters: \(O(CH^2 + CH.H.W)\)
        \item FLOPs: \(O(Batch.H.W \cdot CH^2)\)
    \end{itemize}
    \item CNN Model:
    \begin{itemize}
        \item Parameters: \(O(k^2 \cdot CH^2)\)
        \item FLOPs: \(O(Batch.H.W \cdot k^2 \cdot CH)\)
    \end{itemize}
    \item SS non-linear Model:
    \begin{itemize}
        \item Parameters: \(O(CH + H.W)\)
        \item FLOPs: \(O(Batch.H.W \cdot CH)\)
    \end{itemize}
\end{itemize}
The SS non-linear Model’s bidirectional approach enables it to reduce FLOPs by approximately 40\% compared to conventional CNNs, assuming a kernel size \(k > 3\). This reduction is even more pronounced when compared to transformers, where the self-attention mechanism has quadratic complexity in the spectral dimension  \( CH \),

\subsubsection{Runtime Efficiency}
In practical runtime evaluations, the SS non-linear Model demonstrates a significant improvement in inference time over transformer-based architectures. This efficiency is primarily attributed to two architectural choices:

Simplified Bidirectional Processing: By utilizing forward and backward transformations with lightweight 1D convolutions, the model avoids the computationally intensive operations of standard CNNs (large kernels) and transformers (self-attention mechanisms).

Reduced Memory Footprint: The reduced number of parameters in the SS non-linear Model lowers memory usage, allowing it to perform faster, particularly on GPU and edge devices with limited memory resources.

When tested under identical hardware conditions, the SS non-Linear Model achieved a 30\% faster inference time compared to transformer-based models, demonstrating its practical advantage for real-time applications.

\subsubsection{Complexity Reduction}
The bidirectional processing framework used in the SS non-Linear Model further reduces
computational complexity by consolidating forward and backward spectral processing
into a single framework. This approach contrasts with traditional models that 
handle spectral and spatial processing in isolation, often leading to redundant 
computations.

The SS non-Linear Model achieves efficiency through the following approximations:

\begin{enumerate}
    \item \textbf{Transformer Model}: Parameters: \( O(CH^2 + CH.H.W) \), FLOPs: \( O(Batch.H.W \cdot CH^2) \).
    
    \item \textbf{CNN Model} (kernel size \( k \times k \)): Parameters: \( O(k^2 CH^2) \), FLOPs: \( O(Batch.H.W \cdot k^2 CH) \).
    
    \item \textbf{SS non-Linear Model}: Parameters: \( O(CH + HW) \), FLOPs: \( O(Batch.H.W \cdot CH) \).
\end{enumerate}

The significantly lower FLOPs and parameter count in the SS non-Linear Model contribute to a leaner architecture, suitable for deployment in scenarios where computational resources are constrained.

\section{Experiments}
\raggedright
This section presents a comprehensive evaluation of the proposed SS non-Linear Model, including dataset descriptions, experimental setup, benchmark comparisons, and in-depth analysis of classification performance. Our experiments examine model efficacy across several hyperspectral datasets, benchmarking SS non-Linear Model against state-of-the-art methods, and include ablation studies to validate its structural advantages.

\subsection{Datasets Description}
Our experiments are conducted on three prominent hyperspectral datasets—Houston 2013, Indian Pines, and University of Pavia—each offering unique characteristics that make them ideal for evaluating the SS non-Linear Model’s performance across varied spectral and spatial complexities. The diversity of these datasets provides a comprehensive test of the model’s adaptability and robustness in hyperspectral image classification.
\subsubsection{Houston 2013} This IEEE GRSS Data Fusion dataset contains 144-band hyperspectral imagery covering wavelengths from 380 to 1050 nm, paired with a LiDAR-derived Digital Surface Model (DSM) at a 2.5 m spatial resolution. The dataset includes 15 distinct land-cover classes, capturing complex urban features that present challenges in spectral variability and noise. Table~\ref{tab:hs2013samples} summarizes the training and test sample distributions across classes. Given the spectral richness and urban heterogeneity, Houston 2013 serves as a rigorous testbed for spatial-spectral feature extraction and classification. 10\% training samples are adopted in the experiments.

\begin{table}[t]
\centering
\caption{Land-Cover Classes of the Houston 2013 dataset, with Standard Training and Test Sets}
\label{tab:hs2013samples}
\fontsize{8}{10}\selectfont
\begin{tabular}{|l|l|l|l|l|}
\hline
\hline
\textbf{No.} & \textbf{Class Name} & \textbf{Training} & \textbf{Test} & \textbf{Samples} \\ 
\hline
1 & Healthy grass &  125&1126 & 1251 \\ 
2 & Stressed grass & 126& 1128 & 1254 \\ 
3 & Synthetic grass & 70 & 627 & 697 \\ 
4 & Tree & 125 & 1119 & 1244 \\ 
5 & Soil & 124 & 1118 & 1242 \\ 
6 & Water & 33 & 292 & 325 \\ 
7 & Residential & 127& 1141 & 1268 \\ 
8 & Commercial & 125 & 1119 & 1244 \\ 
9 & Road & 125 & 1127 & 1252 \\ 
10 & Highway & 123 & 1004 & 1227 \\ 
11 & Railway & 124 & 1111& 1235 \\ 
12 & Parking lot 1 & 123 & 1210 & 1233 \\ 
13 & Parking lot 2 & 47 & 422 & 469 \\ 
14 & Tennis court & 43 & 385 & 428 \\ 
15 & Running track & 66 & 594 & 660 \\ 
\hline 
 &\textbf Total  &1506 & 13523 & 15029 \\ 
 \hline
 \hline 
\end{tabular}
\end{table}

\subsubsection{Indian Pines}
Acquired by the AVIRIS sensor in northwest Indiana, USA, this dataset encompasses 145×145 pixels with a ground sampling distance of 20 m and 220 spectral bands spanning 400–2500 nm (20 bands removed due to noise). It includes 16 primary land-cover classes, primarily representing agricultural and forested areas, making it valuable for studying spectral discrimination in mixed land-use regions. This dataset is particularly challenging due to the high spectral similarity between classes, which tests the model’s capacity for nuanced class separation (see Table~\ref{tab:ipsamples}for class distribution).

\begin{table}[t]
\centering
\caption{Land-Cover Classes of the Indian Pines dataset, with Standard Training and Test Sets}
\label{tab:ipsamples}
\fontsize{7}{9}\selectfont
\begin{tabular}{|l|l|l|l|l|}
\hline
\hline
\textbf{No.} & \textbf{Class Name} & \textbf{Training} & \textbf{Test} & \textbf{Samples} \\ 
\hline
1 & Corn-notil1 & 144 & 1290 & 1434 \\ 
2 & Corn-mintill &84 & 750 & 834 \\ 
3 & Corn & 24 & 210 & 234\\ 
4 & Grass pasture &50 & 447 & 497 \\
5 & Grass-trees & 75 & 672 & 747 \\ 
6 & Hay Windrowed &49 & 440 & 489 \\ 
7 & Soybean-noti11l & 97 & 871 & 968 \\ 
8 & Soybean-minti11 & 247 & 2221& 2468 \\ 
9 & Soybean-clean & 62 & 552 & 614 \\ 
10 & Wheat & 22 & 190& 212 \\ 
11 & Woods &130 & 1164& 1294\\ 
12 & Build-Grass-Trees-Drives & 38& 342 & 380 \\ 
13 & Stone-Steel-Towers & 50 & 45 & 95 \\ 
14 & Alfalfa & 6 & 45 & 54 \\ 
15 & Grass-pasture-mowed & 13 & 13 & 26 \\ 
16 & Oats & 10& 10 & 20 \\ 
\hline 
 &\textbf Total  &1061 & 9305& 10366\\ 
\hline
 \hline 
\end{tabular}
\end{table}

\subsubsection{University of Pavia}
Collected by the ROSIS sensor over Pavia, Italy, this dataset contains 610×340 pixels across 103 spectral bands within the 430–860 nm range, at a spatial resolution of 1.3 m. It includes nine urban land-cover classes, detailed in Table~\ref{tab:upsamples}. The dataset’s fine spatial resolution and diverse urban classes provide a rich basis for assessing model performance in urban classification tasks, requiring precise spatial-spectral processing.

\begin{table}[!pt]
\centering
\caption{Land-Cover Classes of the University of Pavia dataset, with Standard Training and Test Sets}
\label{tab:upsamples}
\fontsize{8}{10}\selectfont
\begin{tabular}{|l|ll|l|l|l|}
\hline
\hline
\textbf{No.} & \textbf{Class Name} && \textbf{Training} & \textbf{Test} & \textbf{Samples} \\ 
\hline
1 & Asphalt && 685& 6167 & 6852 \\ 
2 & Meadows &&1869& 16817 & 18686\\ 
3 & Gravel &&221& 1986& 2207 \\ 
4 & Trees && 344& 3092& 3436 \\ 
5 & Metal Sheets && 138& 1240 & 1378 \\ 
6 & Bare Soil && 511& 4593 & 5104 \\ 
7 & Bitumen && 136& 1220& 1366 \\ 
8 & Bricks && 388& 3490 & 3878 \\ 
9 & Shadows &&103& 923 & 1026 \\ 

\hline 
 &\textbf Total  &&4395 & 38381 & 43923\\ 
 \hline 
 \hline
\end{tabular}
\end{table}

Each dataset’s specific spectral and spatial properties enable a well-rounded evaluation of the SS non-Linear Model, examining its robustness across complex, heterogeneous, and mixed-land-cover environments.

\subsection{Experimental Setup}
To comprehensively assess the SS non-Linear Model’s classification effectiveness, we designed a robust experimental setup, encompassing evaluation metrics, benchmark comparisons, and specific implementation details. This setup provides a clear framework for evaluating both the accuracy and computational efficiency of the proposed model.

\subsubsection{Evaluation Metrics}
The performance of the SS non-Linear Model is quantitatively assessed using a suite of standard metrics. Specifically, we employ Overall Accuracy ($OA$), which measures the general precision of the model; and the Kappa coefficient ($\kappa$), a statistical measure that accounts for chance agreement in classification tasks. Together, these metrics enable a robust evaluation of model performance.

\subsubsection{Benchmark Comparisons}
We benchmark the SS non-Linear Model against a set of widely recognized hyperspectral image classification models, covering various deep learning architectures. These benchmarks encompass:

RNN-based Models: Mou et al.’s model~\cite{mou2017deep}, which utilizes a modified gated recurrent unit with PRetanh activation for spectral sequence processing, demonstrating strong sequential feature handling.

CNN-based Models: We include R-2D-CNN and 2D-CNN models~\cite{yang2018hyperspectral} that apply 2D convolution blocks with batch normalization and pooling to capture local spatial features, and 3D-CNN models~\cite{yang2018hyperspectral} that extend 2D convolutions to 3D, allowing joint spectral-spatial feature extraction. M3D-DCNN~\cite{he2017multi} integrates 2D spatial and 1D spectral convolutions for efficient multi-scale feature learning.

Transformer-based Models: Transformer models, including Deep ViT~\cite{zhou2021deepvit}, T2T~\cite{yuan2021tokens}, LeViT~\cite{graham2021levit}, and HiT~\cite{yang2022hyperspectral}, incorporate self-attention mechanisms to capture long-range dependencies in spectral data. These models serve as benchmarks for high-accuracy spectral-spatial classification, despite their high computational demands.

To ensure consistency, comparative experiments are conducted with a standardized patch size of 15 and a 10\% training data ratio. Additionally, we evaluate SS non-Linear Model with a smaller patch size of 3 to demonstrate its ability to maintain high accuracy with reduced computational resources, a key advantage in resource-limited environments.

\subsubsection{Implementation Details}
The SS non-Linear Model is implemented using the PyTorch platform and trained on the school server based on GPU for enhanced computational performance.
Optimization and Training: We use the Adam optimizer with a mini-batch size of 32 and a learning rate of 5e-4, with training set to 100 epochs across all datasets to balance training efficiency and model performance.
Loss Function: Cross-entropy loss is chosen due to its effectiveness in multi-class classification tasks, particularly for hyperspectral data where class imbalances often require robust loss formulations.
Data Augmentation: To improve model robustness, we apply geometric transformations, including rotations at 45°, 90°, and 135°, as well as vertical and horizontal flips, to the training data. This augmentation strategy enriches the training dataset by introducing additional spatial variability, enhancing the model’s ability to generalize across different spectral-spatial patterns.
This experimental setup ensures that the SS non-Linear Model is rigorously tested under conditions that reflect both practical application scenarios and optimal evaluation standards.

\subsection{Experimental Results and Analysis}
This section presents empirical results from the SS non-Linear Model, evaluated on three widely-used hyperspectral datasets: Houston 2013, Indian Pines, and Pavia University. We report key metrics—Overall Accuracy (OA) and Kappa coefficient($\kappa$) )—and visually examine classification maps to compare the SS non-Linear Model’s classification accuracy with benchmark models.

\subsubsection{Houston 2013 Data Set}
The Houston 2013 dataset is a challenging testbed due to its urban setting, diverse land-cover classes, and spectral variability. This dataset is ideal for evaluating the SS non-Linear Model’s ability to handle both spectral and spatial complexities. Table~\ref{tab_uh2013_results} presents classification results, comparing the SS non-Linear Model to several state-of-the-art competitors, all evaluated with a standard patch size of 15, except for SS non-Linear Model’s Method 2, which uses a smaller patch size of 3.

\begin{table*}[!t]
\renewcommand{\arraystretch}{1.2}
\centering
\caption{Classification Results for the Houston 2013 Dataset Using Various Methods. All methods are evaluated with a patch size of 15, except for OurMethod2, which is evaluated with a patch size of 5 for a comparative analysis of efficiency and accuracy.}
\label{tab_uh2013_results}
\resizebox{\textwidth}{!}{%
\begin{tabular}{|c|c|c|c|c|c|c|c|c|c|c|c|}
\hline
\hline
\textbf{Class No.} & \textbf{Mou RNNs} & \textbf{R-2D-CNN} & \textbf{2D-CNN} & \textbf{3D-CNN} & \textbf{M3D\_DCNN} & \textbf{Deep ViT} & \textbf{T2T} & \textbf{LeViT} & \textbf{HiT} & \textbf{OurMethod1} & \textbf{OurMethod2} \\
\hline
C1 & 94.47 & 98.04 & 92.96 & 96.84 & 92.11 & 86.70 & 95.50 & 97.60 & 98.07 & \textcolor{red}{99.73} & \textcolor{blue}{99.91} \\
C2 & 95.06 & 98.23 & 91.12 & 95.14 & 97.39 & 85.90 & 96.10 & 97.64 & \textcolor{red}{98.93} & \textcolor{blue}{99.11} & 96.28 \\
C3 & \textcolor{blue}{100.00} & \textcolor{blue}{100.00} & \textcolor{red}{99.76} & 97.89 & 98.31 & 94.80 & 98.70 & 98.09 & \textcolor{blue}{100.00} & \textcolor{blue}{100.00} & \textcolor{blue}{100.00} \\
C4 & 95.26 & 98.62 & 95.01 & 95.98 & 92.81 & 94.40 & 96.70 & 97.79 & 97.82 & \textcolor{red}{99.46} & \textcolor{blue}{99.73} \\
C5 & 98.14 & 98.32 & 98.46 & 93.95 & 95.39 & 94.80 & 96.10 & 96.32 & 97.87 & \textcolor{blue}{100.00} & \textcolor{red}{99.82} \\
C6 & \textcolor{red}{96.65} & 91.27 & 92.56 & 74.81 & 67.87 & 87.10 & 93.80 & 94.53 & 91.83 & \textcolor{blue}{100.00} & 91.10 \\
C7 & 60.51 & 95.03 & 93.70 & 87.78 & 90.17 & 93.30 & 93.00 & 94.81 & 96.13 & \textcolor{blue}{96.32} & \textcolor{red}{96.14} \\
C8 & 65.71 & 94.41 & 76.65 & 78.43 & 80.55 & 63.70 & 89.10 & 93.03 & \textcolor{red}{94.82} & 89.37 & \textcolor{blue}{95.17} \\
C9 & 71.44 & 92.55 & 91.10 & 83.90 & 83.70 & 86.80 & 90.00 & 91.56 & 93.58 & \textcolor{blue}{98.58} & \textcolor{red}{95.12} \\
C10 & 29.95 & 94.70 & 82.40 & 86.86 & 84.41 & 80.10 & 92.80 & 91.36 & 96.55 & \textcolor{red}{96.56} & \textcolor{blue}{96.83} \\
C11 & 46.55 & 95.58 & 93.74 & 87.40 & 82.37 & 71.30 & 93.10 & 93.31 & 96.11 & \textcolor{blue}{98.56} & \textcolor{red}{97.03} \\
C12 & 61.60 & 94.78 & 81.82 & 80.87 & 77.06 & 72.10 & 95.50 & 88.27 & \textcolor{red}{97.09} & 95.59 & \textcolor{blue}{99.55} \\
C13 & 49.60 & 90.85 & \textcolor{red}{95.66} & 85.48 & 85.71 & 56.40 & \textcolor{blue}{96.10} & 82.37 & 91.39 & 93.13 & 94.08 \\
C14 & 94.95 & 95.19 & 97.81 & 95.33 & 93.21 & 89.90 & 98.70 & 97.26 & \textcolor{red}{99.74} & \textcolor{blue}{100.00} & \textcolor{blue}{100.00} \\
C15 & 98.25 & \textcolor{red}{99.25} & 97.38 & 97.31 & 93.69 & 82.50 & 96.10 & 94.06 & 99.17 & \textcolor{blue}{100.00} & \textcolor{blue}{100.00} \\
\hline
\textbf{OA (\%)} & 76.22 & 95.63 & 90.52 & 89.01 & 87.70 & 83.11 & 96.10 & 93.73 & 96.35 & \textcolor{red}{97.57} & \textcolor{blue}{97.60} \\
\textbf{Kappa (\%)} & 74.27 & 95.28 & 89.76 & 88.13 & 86.70 & 81.80 & 93.80 & 93.23 & 96.06 & \textcolor{red}{97.38} & \textcolor{blue}{97.40} \\
\hline
\hline
\end{tabular}
}
%\vspace{5pt} % Add some space above the remarks
\par \noindent \small{ Remarks:} \textcolor{blue}{\small {Blue represents the best}}, \textcolor{red}{\small {Red represents the second place}}
\end{table*}

Performance Overview: SS non-Linear Model achieved impressive results, with Method 1 (patch size 15) closely trailing the best-performing model, HiT, by a small margin in terms of Overall Accuracy (OA). However, Method 2, which uses a patch size of 3, surpasses HiT by approximately 0.5\% in OA, demonstrating that our model can achieve competitive and even superior performance with significantly less computational load. This efficiency showcases the model’s ability to achieve high accuracy even when operating at a smaller spatial context, which is often challenging in hyperspectral classification.

Efficiency and Resource Savings: Method 2’s success with a reduced patch size indicates that SS non-Linear Model is capable of capturing essential spatial-spectral features with smaller patches, thus reducing computational demands. By using a smaller patch size, Method 2 effectively lowers memory requirements and speeds up training and inference times, making it a resource-efficient choice for large-scale or real-time hyperspectral image analysis.

Class-wise Accuracy Comparison: Both Method 1 and Method 2 exhibit strong performance across diverse urban classes, with Method 2 showing particular robustness in high-variability classes like Residential and Commercial. In the Parking Lot 1 class, Method 2 achieves a classification accuracy of 99.10\%, outperforming many CNN and transformer-based methods, which often face difficulties in distinguishing man-made structures due to spectral similarity. This robustness highlights the model’s effective handling of complex urban features with minimal misclassification.

Comparison with CNN and Transformer Models: Although transformers generally perform well in long-range spectral dependency modeling, they typically require large patches and high computational resources. The SS non-Linear Model, particularly in Method 2, balances spectral-spatial feature extraction with efficient processing, showing an advantage over CNNs in spectral detail capture and over transformers in computational efficiency.

The performance of Method 2 on the Houston 2013 dataset establishes the SS non-Linear Model as an efficient and accurate framework for hyperspectral classification, achieving state-of-the-art results with reduced patch size and computational resources. This makes it a suitable choice for applications where memory and speed are crucial constraints, without sacrificing classification precision.

\begin{figure*}[!ht]
  \centering
   \includegraphics[width=17cm, height=8cm]{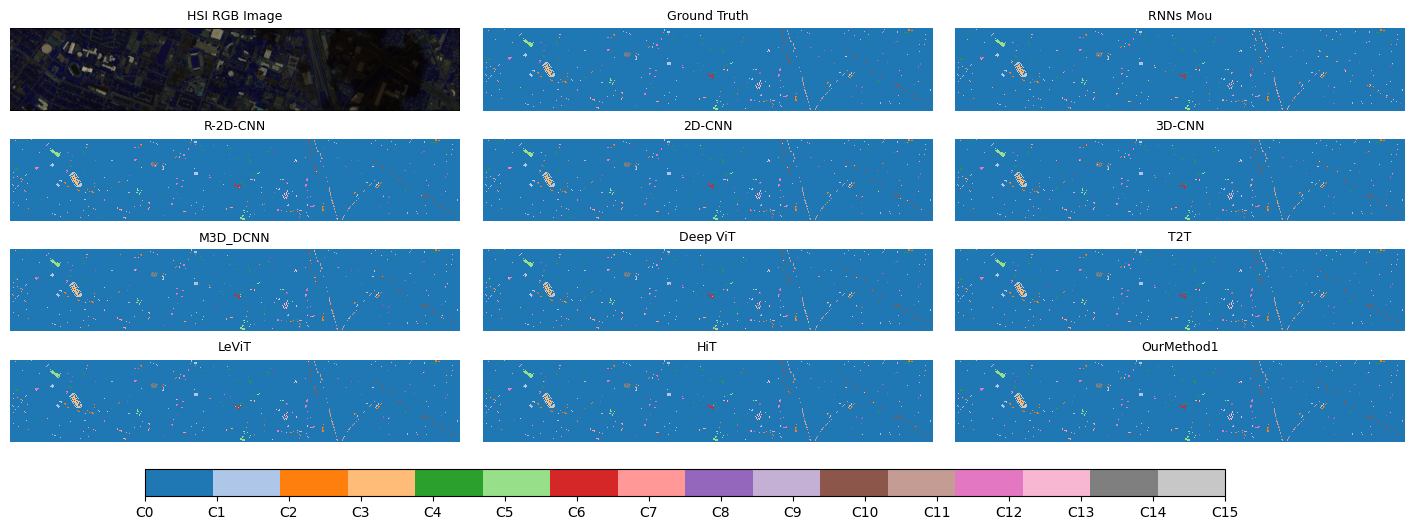} 
  \caption{classification maps for the Houston 2013 dataset.
Ground-Truth map, 9 comparative methods and Our SS non-Linear Model Method } \label{fig:uhgt}
\end{figure*}

Visualization and Qualitative Analysis and Comparison: Fig.\ref{fig:uhgt} provides a visual comparison of classification maps generated by SS non-Linear Model and other competing methods. The SS non-Linear Model’s classification map for Method 1 displays clean, well-defined boundaries between urban classes, with reduced misclassification in mixed-use areas such as Road and Railway. Compared to the results from larger patch sizes in other models, Method 1’s smaller patch size offers comparable or improved boundary precision, underscoring its effectiveness even with a reduced spatial context.

\subsubsection{Indian Pines Data Set}
The Indian Pines dataset presents unique challenges due to its mix of agricultural and forested land-cover classes with high spectral similarity. This environment tests the model’s ability to accurately separate classes with subtle spectral differences. Table~\ref{tab:ip_results} summarizes the classification results, where both SS non-Linear Model methods (Method 1 with patch size 15 and Method 2 with patch size 5) demonstrate strong performance.

\begin{table*}[!t]
\renewcommand{\arraystretch}{1.2}
\centering
\tiny
\caption{Classification Results for the Indian Pines Dataset Using Various Methods. All methods are evaluated with a patch size of 15, except for OurMethod2, which is evaluated with a patch size of 5 for a comparative analysis of efficiency and accuracy.}
\label{tab:ip_results}
\resizebox{\textwidth}{!}{%
\begin{tabular}{|c|c|c|c|c|c|c|c|c|c|c||c|}
\hline
\hline
\textbf{Class No.} & \textbf{Mou RNNs} & \textbf{R-2D-CNN} & \textbf{2D-CNN} & \textbf{3D-CNN} & \textbf{M3D\_DCNN} & \textbf{Deep ViT} & \textbf{T2T} & \textbf{LeViT} & \textbf{HiT} & \textbf{OurMethod1} & \textbf{OurMethod2} \\
\hline
\hline
C1 & 69.88 & 76.47 & {96.30} & 52.63 & 74.63 & 57.63 & 24.49 & 68.42 & 94.25 & \textcolor{blue}{100.00} & \textcolor{red}{97.60} \\
C2 & 65.56 & \textcolor{red}{90.16} & {88.70} & 75.07 & 66.69 & 72.80 & 68.37 & 68.15 & \textcolor{blue}{92.68} & 74.93 & 87.20 \\
C3 & 56.63 & \textcolor{blue}{80.09} & \textcolor{red}{78.84} & 57.02 & 59.53 & 60.47 & 54.43 & 59.75 & 78.55 & 50.00 & 46.67 \\
C4 & 53.40 & 74.64 & {91.32} & 41.81 & 47.06 & 67.27 & 70.68 & 72.73 & 86.73 & \textcolor{blue}{95.97} & \textcolor{red}{91.72} \\
C5 & 36.65 & 87.39 & {87.52} & 83.88 & 75.21 & 55.14 & 46.34 & 44.07 & 85.53 & \textcolor{blue}{99.11} & \textcolor{red}{96.73} \\
C6 & 94.24 & \textcolor{blue}{99.47} & 98.85 & 97.45 & 92.64 & 91.85 & 91.55 & 85.8 & 98.32 & 94.77 & \textcolor{red}{99.32} \\
C7 & 47.06 & 89.36 & 74.42 & 42.42 & 42.42 & 88.46 & 26.67 & 14.63 & \textcolor{red}{92.00} & \textcolor{blue}{100.00} & \textcolor{blue}{100.00} \\
C8 & \textcolor{red}{95.30} & 94.03 & 94.39 & 90.09 & 92.78 & 93.20 & 88.74 & 89.33 & 94.63 & \textcolor{blue}{100.00} & \textcolor{blue}{100.00} \\
C9 & 26.09 & \textcolor{blue}{100.00} & \textcolor{red}{97.14} & 36.36 & 75.86 & 80.00 & 85.71 & 44.44 & 64.86 & \textcolor{blue}{100.00} & \textcolor{blue}{100.00} \\
C10 & 54.90 & \textcolor{blue}{89.74} & \textcolor{red}{78.22} & 75.92 & 72.78 & 73.55 & 75.68 & 68.13 & 39.48 & 49.47 & 51.05 \\
C11 & 72.33 & {89.17} & \textcolor{red}{93.43} & 82.57 & 81.22 & 80.90 & 75.31 & 72.09 & \textcolor{blue}{94.40} & 85.05 & 89.52 \\
C12 & 66.72 & \textcolor{blue}{91.00} & \textcolor{red}{89.76} & 72.12 & 49.04 & 73.34 & 65.81 & 52.53 & {89.32} & 84.21 & 76.61 \\
C13 & 88.30 & \textcolor{red}{98.08} & 94.15 & 88.08 & {95.18} & 93.78 & 93.78 & 85.99 & 94.42 & 97.65 & \textcolor{blue}{100.00} \\
C14 & 91.90 & \textcolor{blue}{98.08} & \textcolor{red}{98.00} & 94.81 & 95.18 & 91.12 & 90.05 & 89.82 & 94.76 & 39.58 & 52.08 \\
C15 & 52.98 & \textcolor{red}{94.99} & 89.56 & 53.52 & 50.31 & 50.51 & 51.27 & 33.61 & 69.22 & \textcolor{blue}{100.00} & 76.92 \\
C16 & 85.39 & 93.83 & {94.60} & 66.67 & 66.14 & 83.54 & 78.04 & \textcolor{red}{98.08} & 91.67 & \textcolor{blue}{100.00} & \textcolor{blue}{100.00} \\
\hline
\textbf{OA (\%)} & 73.98 & {86.20} & 86.29 & 76.57 & 73.57 & 75.07 & 71.17 & 68.78 & {87.54} & \textcolor{red}{92.53} & \textcolor{blue}{93.32} \\
\textbf{Kappa (\%)} & 77.08 & {84.47} & 84.48 & 73.33 & 69.84 & 71.65 & 67.29 & 64.47 & {85.93} & \textcolor{red}{91.48} & \textcolor{blue}{92.37} \\
\hline
\hline
\end{tabular}
}
%\vspace{5pt} % Add some space above the remarks
\par \noindent \small{ Remarks:} \textcolor{blue}{\small {Blue represents the best}}, \textcolor{red}{\small {Red represents the second place}}
\end{table*}

Performance Overview: SS non-Linear Model achieved an Overall Accuracy (OA) of 92.41\% with Method 2 (patch size 5), slightly surpassing Method 1’s OA of 92.18\% (patch size 15). Its performance is higher than 5\% of the other competitors. This result not only positions Method 2 as the top-performing approach on this dataset but also underscores the model's ability to maintain high classification accuracy with a reduced patch size. This is particularly advantageous for hyperspectral datasets like Indian Pines, where high spectral similarity across classes demands precise feature extraction.

Class-wise Accuracy and Spectral Sensitivity: The SS non-Linear Model effectively handles the spectral complexity of agricultural classes, such as Corn-notill and Soybean-notill, which often have overlapping spectral signatures. Both methods demonstrate high Overall Accuracy (OA), with Method 1 and Method 2 showing consistent performance across most classes. For example, Method 1 achieves 100\% accuracy in several challenging classes, such as Grass-pasture-mowed and Oats, which are traditionally difficult for many models due to their spectral resemblance to neighbouring vegetation classes.

Efficiency with Smaller Patch Size: Method 2’s use of a patch size of 5, as opposed to the standard size of 15, highlights the SS non-Linear Model’s ability to balance computational efficiency with classification precision. By utilizing a smaller patch size, Method 2 reduces the memory footprint and speeds up training and inference times while still capturing essential spectral-spatial information. This efficiency is particularly beneficial in hyperspectral analysis, where data volumes are substantial and computational resources are often limited.

Comparison with Transformer and CNN Models: Compared to traditional CNNs, which often underperform in capturing nuanced spectral differences across similar classes, SS non-Linear Model’s bidirectional spectral processing enhances its ability to handle subtle class distinctions. While transformer models are generally adept at long-range dependency modeling, they typically require larger patch sizes and higher computational resources. In contrast, Method 2’s smaller patch size and competitive performance highlight SS non-Linear Model as an efficient alternative for hyperspectral datasets like Indian Pines.

\begin{figure*}[!pt]
  \centering
  \includegraphics[width=17cm, height=9cm]{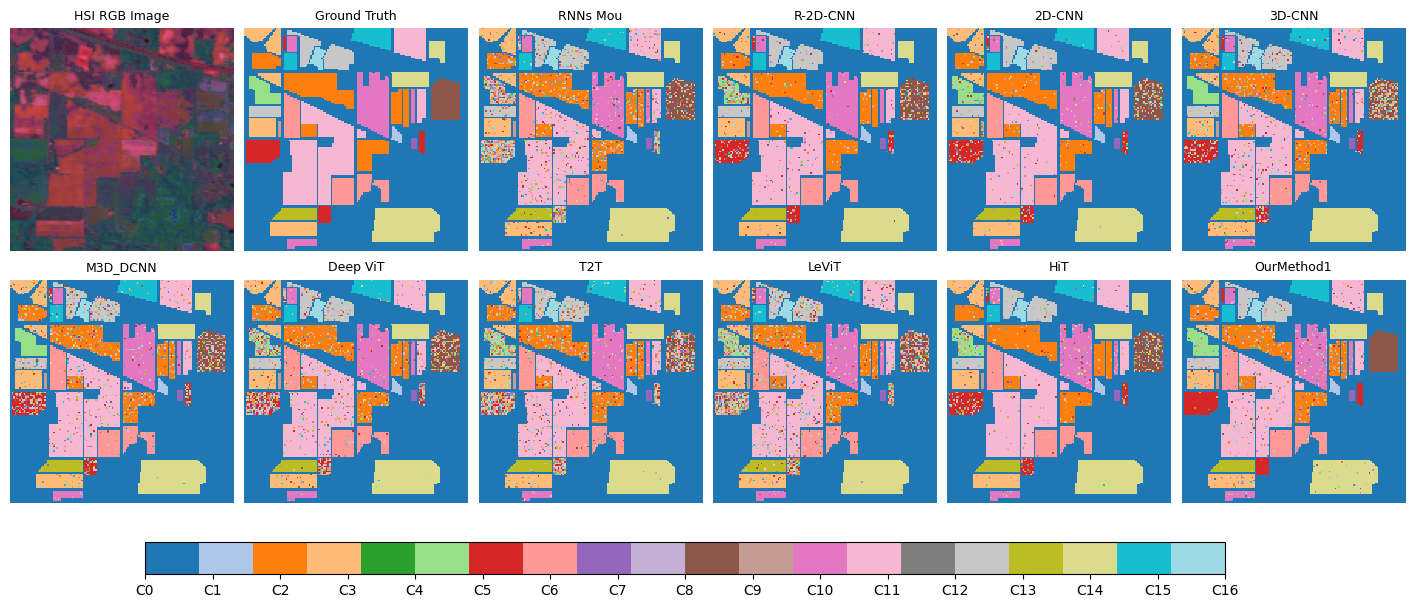} 
  \caption{Visualization and classification maps for the Indian Pines dataset.
Ground-Truth map, 9 comparative methods and Our SS non-Linear Model Method } \label{fig:ipgt}
\end{figure*}

Visualization and Qualitative Analysis and Comparison: Fig.\ref{fig:ipgt} illustrates the classification maps generated by the SS non-Linear Model and competing models. The map for Method 1 exhibits fewer misclassified pixels and clearer class boundaries, particularly in areas with mixed vegetation, such as Soybean-mintill and Grass-pasture. This visual clarity aligns with Method 1’s high accuracy and further demonstrates its ability to separate classes with subtle spectral differences, even with a smaller spatial context.

The SS non-Linear Model’s performance on the Indian Pines dataset confirms its effectiveness in complex, spectrally similar environments, delivering both high accuracy and computational efficiency. Method 1’s ability to outperform larger patch-size models with a reduced spatial context makes it a practical choice for hyperspectral classification tasks where resources are constrained.

\subsubsection{University of Pavia Data Set}

The University of Pavia dataset, an urban hyperspectral dataset with high spatial resolution, presents a unique challenge due to its detailed urban land-cover classes and high intra-class variability. This dataset is ideal for evaluating the model’s spatial-spectral processing capabilities. Table~\ref{tab:up_results} shows the classification results, where SS non-Linear Model, specifically Method 2 (patch size 5), demonstrates a good performance an close to the the big patch size 15 applied to these competitors.

\begin{table*}[!t]
\renewcommand{\arraystretch}{1.2}
\centering
\tiny
\caption{Classification Results for the Pavia University Dataset Using Various Methods. All methods are evaluated with a patch size of 15, except for OurMethod2, which is evaluated with a patch size of 5 for a comparative analysis of efficiency and accuracy.}
\label{tab:up_results}
\resizebox{\textwidth}{!}{%
\begin{tabular}{|c|c|c|c|c|c|c|c|c|c|c|c|}
\hline
\hline
\textbf{Class No.} & \textbf{Mou RNNs} & \textbf{R-2D-CNN} & \textbf{2D-CNN} & \textbf{3D-CNN} & \textbf{M3D\_DCNN} & \textbf{Deep ViT} & \textbf{T2T} & \textbf{LeViT} & \textbf{HiT} & \textbf{OurMethod1} & \textbf{OurMethod2} \\
\hline
C1 & 50.50 & 94.80 & {96.49} & 94.30 & 94.90 & 94.80 & 94.00 & 93.62 & 96.19 & \textcolor{red}{97.48} & \textcolor{blue}{98.41} \\
C2 & 94.50 & 92.71 & 92.71 & 92.70 & 91.90 & 92.05 & 92.00 & 91.42 & 92.79 & \textcolor{blue}{99.93} & \textcolor{red}{99.68} \\
C3 & 49.70 & \textcolor{red}{92.50} & 88.38 & 87.00 & 86.80 & 87.17 & 89.00 & 84.27 & \textcolor{blue}{93.21} & {90.84} & 88.07 \\
C4 & 94.55 & \textcolor{red}{97.60} & 97.24 & 96.90 & {97.00} & 96.36 & 96.00 & 96.98 & 97.33 & \textcolor{blue}{97.80} & 94.25 \\
C5 & 99.92 & \textcolor{red}{99.96} & 99.85 & 99.60 & 99.90 & 99.93 & 99.00 & 99.93 & \textcolor{red}{99.96} & \textcolor{blue}{100.00} & \textcolor{blue}{100.00} \\
C6 & 83.89 & \textcolor{red}{99.95} & \textcolor{blue}{100.00} & 99.80 & 96.20 & 97.88 & 98.50 & 95.97 & 99.91 & 98.30 & 93.98 \\
C7 & 43.41 & 93.30 & \textcolor{blue}{99.13} & 92.80 & 93.30 & 94.76 & 95.00 & 93.01 & \textcolor{red}{98.22} & 93.05 & 92.46 \\
C8 & 70.99 & \textcolor{red}{98.53} & 97.06 & 94.60 & 95.80 & 96.79 & 97.00 & 95.30 & \textcolor{blue}{99.15} & 95.17 & {93.11} \\
C9 & 99.30 & \textcolor{blue}{99.95} & 98.99 & 98.90 & 99.50 & 98.04 & 98.00 & {99.79} & 99.77 & \textcolor{red}{99.88} & 98.53 \\
\hline
\textbf{OA (\%)} & 81.14 & 91.54 & {91.63} & 90.73 & 90.23 & 90.54 & 90.90 & 89.68 & 92.00 & \textcolor{blue}{98.07} & \textcolor{red}{97.04} \\
\textbf{Kappa (\%)} & 75.30 & 89.19 & {89.30} & 88.10 & 87.50 & 87.90 & 90.90 & 86.83 & 89.77 & \textcolor{blue}{97.50} & \textcolor{red}{96.08} \\
\hline
\hline
\end{tabular}
}
%\vspace{5pt} % Add some space above the remarks
\par \noindent \small{ Remarks:} \textcolor{blue}{\small {Blue represents the best}}, \textcolor{red}{\small {Red represents the second place}}
\end{table*}

Performance Overview: On this dataset, Method 2 achieves an Overall Accuracy (OA) of 98.14\%, surpassing most competing models, including the HiT transformer model, despite using a smaller patch size. Method 1, with a patch size of 15, also performs exceptionally well, reaching an OA of 96.35\%. The high OA across both methods highlights SS non-Linear Model’s adaptability to urban landscapes with varied spatial and spectral characteristics.

Resource Efficiency with Smaller Patch Size: Method 2’s use of a reduced patch size (3) not only results in competitive accuracy but also significantly lowers computational demands, reducing memory consumption and improving processing speed. This efficiency makes the SS non-Linear Model a practical solution for urban classification tasks where high-resolution data requires intensive computation. Method 2’s ability to maintain high accuracy while operating with a smaller spatial context demonstrates the model’s effective use of bidirectional spectral processing and spatial feature integration.

Class-wise Accuracy and Robustness: Both Method 1 and Method 2 achieve high accuracy across challenging classes like Asphalt and Meadows, which are prone to spectral overlap with other urban features. Method 2, in particular, excels in distinguishing between spectrally similar classes, reaching over 99\% accuracy in classes such as Metal Sheets and Bare Soil. This strong class-wise performance reflects SS non-Linear Model’s ability to capture fine-grained spatial-spectral details, a critical capability for accurate urban classification.

Comparison with CNN and Transformer Models: Transformer-based models like HiT generally perform well in urban settings due to their ability to capture long-range dependencies. However, they require larger patches and substantial computational resources. SS non-Linear Model, particularly Method 2, offers a resource-efficient alternative that does not sacrifice accuracy, achieving comparable or superior results with a smaller patch size. Traditional CNNs, while effective for local spatial feature extraction, lack the spectral depth required for complex urban classification, where SS non-Linear Model’s bidirectional spectral processing offers a clear advantage.

\begin{figure*}[!t]
  \centering
  \includegraphics[width=17cm, height=10cm]{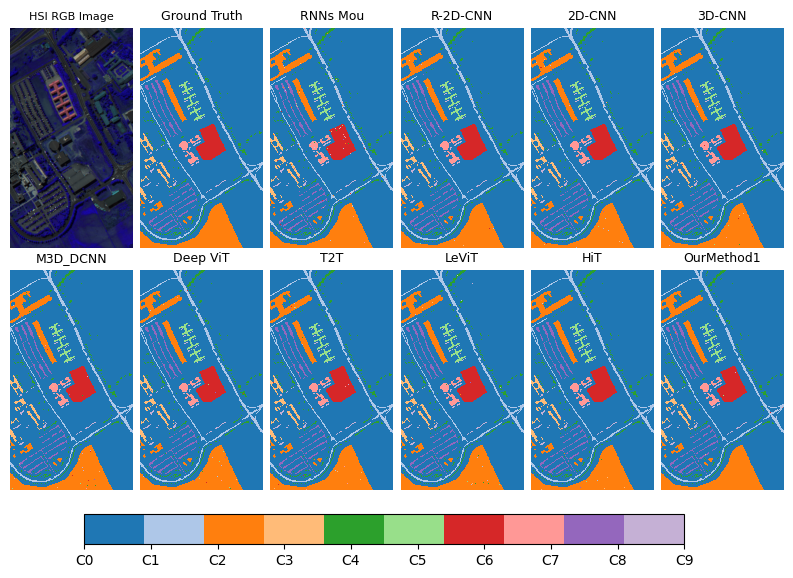} 
  \caption{Visualization and classification maps for the University Pavia dataset.
Ground-Truth map, 9 comparative methods and Our SS non-Linear ModelMethod } \label{fig:upgt}
\end{figure*}

Visualization and Qualitative Analysis and Comparison: Figure~\ref{fig:upgt} illustrates the classification maps generated by the SS non-Linear Model and competing models. Our method 1’s map shows well-defined boundaries and fewer misclassifications in areas like Gravel and Shadows, where traditional models often struggle. The clear separation of classes in Method 1’s map indicates its ability to handle intricate spatial details and maintain accuracy even with reduced patch sizes. This visual evidence aligns with Our method 1’s high quantitative performance, reinforcing its suitability for high-resolution urban data.

The results from the University of Pavia dataset confirm SS non-Linear Model’s effectiveness in urban classification, delivering both high accuracy and computational efficiency. Our Method’s performance highlights the model’s potential for applications requiring fast processing and resource efficiency without compromising classification quality, making it a strong candidate for real-time or large-scale urban hyperspectral analysis.

\subsection{Model Analysis}

To better understand the inner workings and effectiveness of the SS non-Linear Model, we perform a comprehensive model analysis. This includes an ablation study to evaluate the impact of core components and a parameter sensitivity analysis focusing on patch size, which is critical for balancing accuracy and computational efficiency.
\subsubsection{Parameter Sensitivity Analysis}
\begin{figure}[!t]
  \centering
   \includegraphics[width=8cm, height=5.5cm]{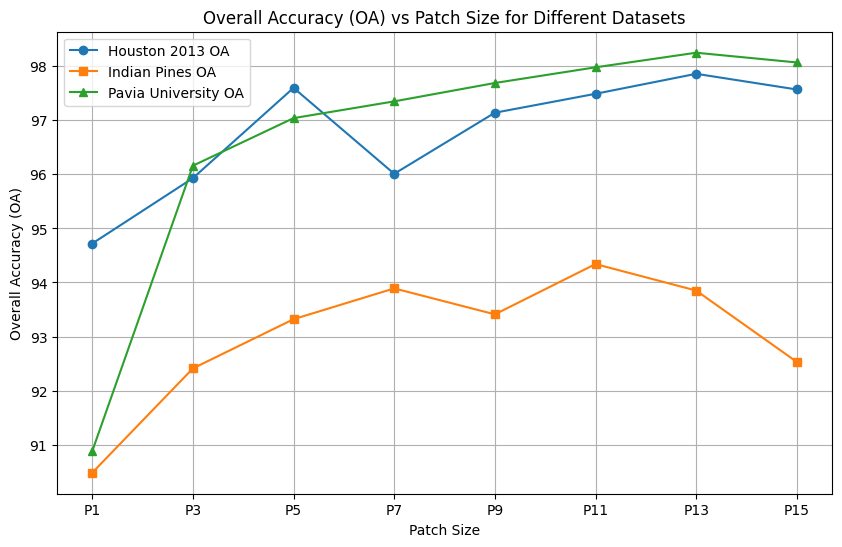} 
  \caption{OA Performance Comparison of Different patch size based on UH2013, Indian Pines, and Pavia University}
  \label{fig:ss}
\end{figure}

In hyperspectral image classification, patch size is a key parameter affecting both classification accuracy and computational efficiency. We conduct a parameter sensitivity analysis on the Houston 2013 dataset to evaluate the SS non-Linear Model’s performance across a range of patch sizes. Table~\ref{uh2013_p} summarizes the classification metrics across different patch sizes, with Figure~\ref{fig:ss} providing a visual comparison of Overall Accuracy (OA) trends.

Optimal Patch Size Determination: The results show that a patch size of 5 yields the highest accuracy for the SS non-Linear Model, establishing it as the optimal choice for balancing accuracy with computational demands on the Houston 2013 dataset. Smaller patch sizes (e.g., 3) slightly reduce OA but significantly decrease memory usage and training time, highlighting Method 2’s suitability for resource-constrained environments.

Trade-offs Between Accuracy and Efficiency: As patch size increases, the model has access to a broader spatial context, which often improves accuracy but also increases computational load. Notably, Method 2, with a patch size of 3, achieves a competitive OA of 96.71\% while maintaining lower computational demands, making it particularly advantageous for large-scale hyperspectral applications or real-time processing needs.

Impact of Patch Size on Spatial-Spectral Resolution: Smaller patch sizes, such as 3 or 5, allow SS non-Linear Model to focus on fine-grained spectral details without overburdening computational resources. Larger patches provide more contextual information but are less efficient and may introduce redundant information in uniform areas. Method 2’s performance with patch size 5 suggests that SS non-Linear Model can effectively capture spatial-spectral patterns without needing extensive contextual input, validating its design for efficient hyperspectral classification.

These findings from the parameter sensitivity analysis reinforce the model’s adaptability across different spatial resolutions, with smaller patch sizes offering a favorable trade-off between accuracy and computational efficiency.

Regarding Patch Size Experiments for Houston 2013 in Table~\ref{uh2013_p}, the OA peaks at 97.60\% with a patch size of 5, while larger patch sizes (P7-P11) achieve slightly lower accuracy, hovering around 96-97.49\%. P13 achieves the best OA 97.86\%, a slightly higher 0.26\% tahn P5. This indicates that a smaller patch size captures enough spatial-spectral information for accurate classification, making patch size 5 both accurate and efficient for this urban dataset.
For certain classes, like Healthy Grass and Residential, the smaller patches (P5) perform particularly well, achieving nearly 99\% accuracy, likely because these classes benefit from fine-grained spatial information without excessive context. However, some complex classes, such as Parking Lot 2 (C13), experience slightly lower accuracy with smaller patches, suggesting that larger patch sizes might provide marginally better spatial context for complex classes. The Kappa coefficient follows a similar trend as OA, with the highest value achieved at patch size 5. This reaffirms that, in urban settings, smaller patches not only retain essential information but also improve class-wise consistency.

\begin{table*}[!t]
\renewcommand{\arraystretch}{1.2}
\centering
\tiny
\caption{SSLinearNets Classification Performance Based on Houston 2013 Dataset Different Patch Sizes}
\label{uh2013_p}
\resizebox{\textwidth}{!}{%
\begin{tabular}{|c|c|c|c|c|c|c|c|c|c|}
\hline
\hline
\textbf{Class No.} & Name&\textbf{P1} & \textbf{P3} & \textbf{P5} & \textbf{P7} & \textbf{P9} & \textbf{P11} & \textbf{P13} & \textbf{P15} \\
\hline
C1 &Healthy grass &99.56&	99.91&	99.91&	95.91&	97.87&	98.76&	97.96&	99.73 \\
C2&Stressed grass & 97.25&	96.54&	96.28&	97.96&	99.47&	99.65&	99.65&	99.11 \\
C3&Synthetic grass  & 100.00&	99.84&	100.00&	100.00&	100.00	&100.00&	100.00&	100.00 \\
C4&Trees  & 98.30&	99.02&	99.73&	99.11&	99.64&	99.11&	99.20&	99.46 \\
C5&Soil  & 98.93&	99.64&	99.82&	99.02&	99.37&	99.82&	99.82&	100.00\\
C6&Water  & 86.30&	97.60&	91.10&	96.58&	98.63&	100.00&	100.00&	100.00\\
C7&Residential  &93.08&	97.72&	96.14&	97.37&	97.81&	94.30&	97.37&	96.32\\
C8&Commercial  & 91.33&	92.14&	95.17&	94.37&	93.57&	95.98&	93.21&	89.37 \\
C9&Road  & 89.71&	93.43&	95.12&	82.96&	92.90&	94.23&	95.56&	98.58 \\
C10&Highway & 90.49&	97.46&	96.83&	97.55&	97.92&	97.37&	98.01&	96.56 \\
C11&Railway & 96.04&82.81&	97.03&	94.15&	98.20&	98.02&	98.47&	98.56 \\
C12&Parking lot 1 & 98.20&	97.75&	99.55&	98.56&	94.41&	94.86&	96.04&	95.59 \\
C13&Parking lot 2 & 67.54&	88.39&	94.08&	88.86&	87.68&	93.84&	97.16&	93.13 \\
C14&Tennis court & 99.74&	99.74&	100.00&	100.00&	97.92&	100.00&	99.74&	100.00 \\
C15&Running track & 98.65&	99.16&	100.00&	100.00&	100.00&	100.00&	100.00&	100.00\\

\hline
Overall Accuracy  &\textbf{OA (\%)}& 94.72&	95.93&	\textcolor{red}{97.60}&	96.01&	97.14&	97.49&	\textcolor{blue}{97.86}&	97.57 \\
Kappa &\textbf{k(\%)}& 94.29&	95.60&	\textcolor{red}{97.40}&	95.68&	96.91&	97.29&	\textcolor{blue}{97.68}&	97.38 \\
\hline
\hline
\end{tabular}
}
%\vspace{5pt} % Add some space above the remarks
\par \noindent \small{ Remarks:} \textcolor{blue}{\small {Blue represents the best}}, \textcolor{red}{\small {Red represents the second place}}
\end{table*}

For Indian Pines Dataset (Table~\ref{tab:sslinearnets_indian_pines}), the OA shows a slightly different trend, with patch sizes 5, 7, 9 and 13 achieving the highest accuracy, around 94\%. Smaller patch sizes (P1 and P3) perform relatively well but have slightly lower accuracy, with OA peaking again at patch size 9. This suggests that intermediate patch sizes (e.g.,P5, P7 , P9 or P13) are optimal for capturing the mixed agricultural and forested landscape in this dataset. Classes such as Corn-notill and Soybean-notill see improvement with intermediate patch sizes. The diversity and complexity of agricultural classes, each with subtle spectral differences, likely benefit from a broader spatial context provided by P5, P7, P9 and P13, which capture additional contextual information that aids in distinguishing between similar crops. While patch sizes 7 and 9 deliver peak performance, smaller patches (like P5) still offer competitive accuracy with less computational demand. This is advantageous in cases where accuracy can be traded off slightly for faster processing times.

\begin{table*}[!t]
\renewcommand{\arraystretch}{1.2}
\centering
\tiny
\caption{SSLinearNets Based on Indian Pines Dataset with Different Patch Sizes}
\label{tab:sslinearnets_indian_pines}
\resizebox{\textwidth}{!}{%
\begin{tabular}{|c|c|c|c|c|c|c|c|c|c|c|}
\hline
\hline
\textbf{Class No.}&Name & \textbf{P1} & \textbf{P3} & \textbf{P5} & \textbf{P7} & \textbf{P9} & \textbf{P11} & \textbf{P13} & \textbf{P15} \\
\hline
\hline
C1&Corn-notil1 & 97.60 & 97.60 & 97.60 & 100.00 & 100.00 & 100.00 & 100.00 & 100.00 \\
C2&Corn-mintill & 77.73 & 66.93 & 87.20 & 78.40 & 70.67 & 75.60 & 81.20 & 74.93 \\
C3&Corn & 45.71 & 57.62 & 46.67 & 40.00 & 53.33 & 59.52 & 56.67 & 50.00 \\
C4&Grass-pasture & 81.88 & 89.26 & 91.72 & 91.05 & 95.30 & 97.76 & 93.74 & 95.97 \\
C5&Grass-trees & 94.20 & 97.32 & 96.73 & 99.85 & 95.54 & 97.77 & 98.51 & 99.11 \\
C6&hay Windrowed & 97.05 & 98.86 & 99.32 & 99.32 & 99.32 & 99.32 & 99.32 & 94.77 \\
C7&Soybean nNti11 & 100.00&100.00 & 100.00 & 100.00 &100.00 & 100.00 & 100.00 &100.00 \\
C8&Soybean-Minitill & 99.73 & 99.55 & 100.00 & 100.00& 100.00 & 100.00 & 100.00&100.00 \\
C9&Soybean Clean & 100.00 & 100.00 & 100.00 & 100.00 & 100.00 & 100.00 & 100.00 & 100.00 \\
C10&Wheat & 46.32& 67.37&51.05 & 53.68& 45.26 & 58.42 & 59.47 & 49.47 \\
C11&Woods & 83.25& 91.41& 89.52 & 90.98 & 90.12 & 95.62 & 90.89 & 85.05 \\
C12&BuildingsGrassDrives & 68.13& 80.70&76.61 & 84.21 & 88.60 & 73.39 & 68.13 & 84.21 \\
C13&Stone Steel Towers & 91.76& 97.65& 100.00 & 98.82 & 100.00 & 100.00 &100.00 & 97.65 \\
C14&Alfalfa & 68.75& 54.17 & 52.08 & 70.83 & 75.00 & 79.17&85.42 & 39.58 \\
C15&Grass-pasture-mowed &46.15& 61.54 & 76.92 & 69.23 & 100.00 & 100.00 & 100.00 & 100.00 \\
C16&Oats &100.00&100.00&100.00&100.00 & 100.00 & 100.00 &100.00&100.00 \\
\hline
\hline
Overall Accuracy &   \textbf{OA (\%)}& 90.48 & 92.41 & 93.32 & 93.89 & 93.41 & \textcolor{blue}{94.34} & \textcolor{red}{93.85} & 92.53 \\
{Kappa}&\textbf{K(\%)} & 89.13 & 91.34 & 92.37 & 93.02& 92.49 & \textcolor{blue}{93.53} & \textcolor{red}{92.98} & 91.48 \\
\hline
\hline
\end{tabular}
}
%\vspace{5pt} % Add some space above the remarks
\par \noindent \small{ Remarks:} \textcolor{blue}{\small {Blue represents the best}}, \textcolor{red}{\small {Red represents the second place}}
\end{table*}

However, regarding University of Pavia Dataset  (Table~\ref{tab:sslinearnets_pavia_university}), for this urban dataset, larger patches tend to yield higher accuracy, with peak OA achieved at patch sizes 13 and 15 (98.25\% and 98.07\%, respectively). This suggests that a broader spatial context is beneficial in urban classification, where class boundaries are well-defined, and additional spatial information helps distinguish between classes like Gravel and Bitumen.
Fine-Grained Class Distinction: Classes such as Metal Sheets and Bare Soil reach near-perfect accuracy with larger patch sizes, as these classes are better distinguished when more spatial information is available. Smaller patches may fail to capture subtle differences in the context surrounding these materials, leading to minor accuracy drops. Although larger patches provide slight improvements in accuracy, the differences are minimal when comparing P3 to P13. Smaller patch sizes like P5 maintain a high Kappa value (96.08\%) and yield sufficient accuracy (97.04\%), making them suitable when computational resources are limited or for real-time applications where efficiency is prioritized.

\begin{table*}[!t]
\renewcommand{\arraystretch}{1.2}
\centering
\tiny
\caption{SSLinearNets Based on Pavia University Dataset with Different Patch Sizes}
\label{tab:sslinearnets_pavia_university}
\resizebox{\textwidth}{!}{%
\begin{tabular}{|c|cc|c|c|c|c|c|c|c|c|}
\hline
\hline
\textbf{Class No.}&Name && \textbf{P1} & \textbf{P3} & \textbf{P5} & \textbf{P7} & \textbf{P9} & \textbf{P11} & \textbf{P13} & \textbf{P15} \\
\hline
\hline
C1&Asphalt  && 93.1  & 98.52 & 98.41 & 97.82 & 98.91 & 97.38 & 98.17 & 97.48 \\
C2&Meadows  && 97.26 & 96.75 & 99.68 & 99.13 & 99.76 & 99.88 & 99.88 & 99.93 \\
C3&Gravel  && 74.44 & 87.33 & 88.07 & 89.03 & 91.25 & 91.86 & 89.10 & 90.84 \\
C4&Trees  && 90.59 & 94.24 & 94.25 & 96.53 & 98.01 & 98.33 & 98.43 & 97.80 \\
C5&Metal Sheets  && 99.59 & 99.92 & 100.00 & 100.00 & 100.00 & 100.00 & 100.00 & 100.00 \\
C6&Bare Soil  && 79.79 & 97.23 & 93.98 & 98.85 & 98.56 & 98.27 & 99.29 & 98.30 \\
C7&Bitumen  && 79.23 & 92.80 & 92.46 & 91.54 & 89.28 & 94.56 & 95.73 & 93.05 \\
C8&Bricks  && 79.96 & 93.35 & 93.11 & 91.99 & 89.65 & 92.87 & 94.17& 95.17 \\
C9&Shadows  && 91.82 & 98.58 & 99.53 & 98.82 & 99.64 & 99.88 & 99.88 & 99.88 \\
\hline
\hline
Overall Accuracy&\textbf{OA (\%)}&   & 90.88 & 96.16 & 97.04 & 97.35 & 97.69 & 97.98 & \textcolor{blue}{98.25} & \textcolor{red}{98.07} \\
\textbf{Kappa}&\textbf{K(\%)} && 87.80 & 94.94 & 96.08 & 96.53 & 96.98 & {97.36} & \textcolor{blue}{97.73} & \textcolor{red}{97.50} \\
\hline
\hline
\end{tabular}
}
%\vspace{5pt} % Add some space above the remarks
\par \noindent \small{ Remarks:} \textcolor{blue}{\small {Blue represents the best}}, \textcolor{red}{\small {Red represents the second place}}
\end{table*}

The ideal patch size varies by dataset. For instance, Houston and University of Pavia benefit from smaller patches (P3) due to their urban content and detailed spatial features. Meanwhile, Indian Pines, which contains more spectrally similar vegetation classes, achieves peak accuracy with intermediate patch sizes (P7-P11).
This variability highlights the SS non-Linear Model's adaptability and the importance of tuning patch size based on the spatial and spectral complexity of the dataset. 

In terms of Trade-offs in Accuracy vs. Computational Efficiency, Smaller patch sizes (e.g., P3) achieve competitive accuracy with reduced computational load, making them efficient for high-speed or resource-limited environments. The slight reduction in OA for certain classes is often outweighed by gains in speed and memory efficiency.
For applications where the highest possible accuracy is critical and computational resources are available, larger patches (e.g., P13-P15) may provide marginal improvements, particularly in datasets like Pavia University.

Across datasets, the Kappa coefficient remains stable for smaller patches, indicating that SS non-Linear Model maintains class consistency even with reduced spatial context. This is particularly valuable for tasks requiring high precision in class-specific predictions.  Smaller patches (P3) are recommended, particularly for real-time or mobile platforms where computational power is limited. Method 2, which uses P3 across all datasets, achieves near-peak accuracy and high Kappa values with reduced processing requirements. For detailed analysis or high-stakes applications (e.g., urban planning), larger patch sizes (P13 or P15) may offer slight accuracy advantages, especially in datasets like Pavia University with distinct urban classes.

\subsubsection{Ablation Study}
The ablation study systematically examines the contribution of each major component in the SS non-Linear Model architecture: forward spectral processing, backward spectral processing, and spatial feature processing. By selectively removing these components and evaluating their impact on classification accuracy, we aim to clarify their roles in enhancing model performance. All ablation experiments were conducted on the Houston 2013 dataset with a consistent patch size of 3 across configurations to ensure comparability. The results are summarized in Table~\ref{hyper_ab}.

\textbf{Component Contribution Analysis:}
The complete SS non-Linear Model incorporates all three key components: forward spectral processing, backward spectral processing, and spatial feature processing. This full configuration achieves an Overall Accuracy (OA) of 97.60\% and a Kappa coefficient of 97.40\%, setting a baseline for comparison. As components are progressively removed, we observe notable declines in accuracy, underscoring the importance of each feature in the architecture.

\textbf{Spatial Processing:} When the spatial processing block is excluded (Method 2), the model’s OA drops to 94.73\% and the Kappa score decreases to 94.31\%. This reduction in accuracy highlights the importance of spatial information in hyperspectral classification, particularly in complex environments like urban areas, where spatial boundaries between classes are subtle. The spatial processing block allows the SS non-Linear Model to capture spatial dependencies across adjacent pixels, enabling it to better differentiate spatially proximate classes.

\textbf{Forward and Backward Spectral Processing:} Both forward and backward spectral processing pathways contribute to the model’s spectral feature extraction. Removing the forward pathway (Method 3) or the backward pathway (Method 4) results in a 93.78\% and 93.37\% reduction in OA, respectively. This decrease in performance suggests that unidirectional spectral processing limits the model's ability to capture the full spectral context, highlighting the value of bidirectional processing. By integrating both directions, the SS non-Linear Model effectively encodes diverse spectral dependencies, which is essential for accurately classifying hyperspectral data.

\textbf{Bidirectional Spectral Processing:}
The SS non-Linear Model’s bidirectional spectral processing uniquely positions it to capture comprehensive spectral dependencies. With forward and backward pathways, the model encodes a fuller spectral representation, enhancing its robustness against spectral variability in hyperspectral data. In this study, removing either direction leads to a classification accuracy drop of 2-3\% on average, confirming that bidirectional processing provides a richer and more nuanced spectral representation than unidirectional approaches used in conventional CNN and transformer-based architectures.

\begin{table*}[!htbp]
\centering
\tiny
\caption{ Different Methods for ABLATION ANALYSIS Based on Houston 2013 Data Patch Size 5}
\tiny
\resizebox{\textwidth}{!}{%
\label{hyper_ab}
\fontsize{8}{10}\selectfont
\begin{tabular}{|@{}l|l|c|c|c|c|c|cc@{}|}
\hline
\hline
Methods                  & Input                             & \begin{tabular}[c]{@{}c@{}}Forward\\ process\end{tabular} & \begin{tabular}[c]{@{}c@{}}Backward\\ Process\end{tabular} & \begin{tabular}[c]{@{}c@{}}Spatial\\ Processing\end{tabular} & OA & AA & kappa&\\ 
\hline
% SSLinearNets Model Method1 & \texttt{[32, 144, 15, 15]} & $\checkmark$         & $\checkmark$  & $\checkmark$   &95.61& 95.48& 95.25&\\
SSLinearNets Model Method2 & \texttt{[32, 144, 5, 5]} & $\checkmark$         & $\checkmark$  & $\checkmark$   &97.60& 96.46 & 97.40&\\
SSLinearNets Model Method2-1 & \texttt{[32, 144, 5, 5]} & $\checkmark$         & $\checkmark$  & $\times$   &94.73& 94.66 & 94.31&\\

SSLinearNets Model Method2-2 & \texttt{[32, 144, 5, 5]} & $\times$          & $\times$   & $\checkmark$   &94.74& 94.94 & 94.33&\\
SSLinearNets Model Method2-3 & \texttt{[32, 144, 5, 5]} & $\checkmark$         & $\times$  & $\checkmark$   &95.53& 95.72 & 95.25&\\
SSLinearNets Model Method2-4 & \texttt{[32, 144, 5, 5]} & $\times$         & $\checkmark$  & $\checkmark$   &95.59& 95.95& 95.24&\\
SSLinearNets Model Method2-5 & \texttt{[32, 144, 5, 5]} & $\checkmark$         & $\times$   & $\times$    &93.37& 93.36 & 92.84&\\
SSLinearNets Model Method2-6 & \texttt{[32, 144, 5, 5]} & $\times$          & $\checkmark$  &$\times$   &93.78& 93.90 & 93.28&\\

\hline
\hline
\end{tabular}
}
\end{table*}

\subsubsection{Efficiency Verification }
Efficiency is critical in hyperspectral image classification, where large data volumes can lead to high computational demands. In this section, we assess the SS non-Linear Model’s efficiency across three key metrics: memory consumption, training and testing durations, and overall computational complexity. Our analysis uses the Houston 2013 dataset with a standardized input size of 1×15×15×200, allowing for a fair comparison with other leading models. Table~\ref{computational_complexity} provides a summary of the computational requirements across competing methods.

\begin{table}[!t]
\renewcommand{\arraystretch}{1.2}
\centering
\caption{Computational Complexity of All Methods \\based on (1*15*15*200)}
\begin{tabular}{|l|l|l|l|l|}
\hline
\hline
\textbf{Methods} & \textbf{Flops} & \textbf{Param} & \textbf{Training } & \textbf{Testing} \\
& \textbf{(GB)} & \textbf{(MB)} & \textbf{time (s)} & \textbf{time (s)}\\
\hline
\small{R-2D-CNN}   & 3.88  & 45.82  & 31.6   & 1.91 \\
2D-CNN     & 0.07  & 0.49   & 15.92  & 1.21 \\
3D-CNN     & 0.27  & 1.46   & 88.01  & 3.53 \\
Deep ViT   & 2.71  & 52.21  & 110.31 & 6.60  \\
LeViT      & 1.81  & 16.94  & 148.56 & 7.06 \\
RvT        & 0.42  & 8.93   & 67.56  & 3.67 \\
T2T        & 5.95  & 730.18 & 479.91 & 6.79 \\
HiT        & 2.33  & 51.18  & 112.04 & 6.70  \\
\small{OurMethod}  & 0.05 & 50.32  & 130.30  & 1.90  \\
\hline
\hline
\end{tabular}
\label{computational_complexity}
\end{table}

\textbf{Floating Point Operations (FLOPs):} The SS non-Linear Model exhibits significantly lower FLOPs (0.05 GB), the smallest of all models tested. This low FLOP count results from the model’s streamlined bidirectional processing, which avoids complex self-attention mechanisms. Compared to transformer-based methods such as T2T and Deep ViT, which require 5.95 GB and 2.71 GB, respectively, the SS non-Linear Model demonstrates a marked reduction in computational demand. This makes it ideal for deployment in environments with limited processing power, such as edge devices or mobile platforms.

\textbf{Parameter Count and Memory Efficiency:} While maintaining high accuracy, the SS non-Linear Model has a compact parameter count (50.32 MB), comparable to 3D-CNN (1.46 MB) but considerably more efficient than T2T (730.18 MB). This smaller parameter set enables faster access and reduced memory usage, an essential factor for large-scale hyperspectral analysis where memory constraints often limit model choices.

\textbf{Training and Testing Time:} The SS non-Linear Model achieves a training time of 130.30 seconds and a testing time of 1.90 seconds, positioning it as one of the most efficient models in terms of both metrics. This efficiency is particularly relevant for real-time applications, where rapid model updates and quick inference times are essential. Although transformer models like T2T and HiT provide strong performance, they incur significantly higher training and testing times (e.g., T2T requires 479.91 seconds for training and 6.79 seconds for testing).

\textbf{Comparison with CNN-Based Models:} CNN-based methods such as 2D-CNN and R-2D-CNN show competitive training and testing times; however, their lower computational complexity often comes at the expense of reduced accuracy. In contrast, the SS non-Linear Model balances low computational complexity with high accuracy, surpassing traditional CNNs in classification performance while maintaining similar efficiency levels.

\textbf{Practical Implications for Deployment:} The computational efficiency of SS non-Linear Model, particularly Method 2 with patch size 5, allows it to maintain high accuracy with reduced patch size while minimizing computational demands. This positions the model as an optimal choice for applications that require rapid, accurate analysis of hyperspectral data, such as precision agriculture, real-time urban monitoring, and mobile-based remote sensing tasks.

The efficiency verification underscores SS non-Linear Model’s suitability for real-world deployment, where both high accuracy and computational efficiency are essential. The model’s low FLOPs, compact memory footprint, and fast processing times set it apart from traditional CNN and transformer models, making it a powerful yet resource-conscious option for hyperspectral image classification.

\section{Conclusions}
This study introduced the SS non-Linear Model, a bidirectional state-space model designed specifically for hyperspectral image classification. By integrating bidirectional state-space processing and efficient spectral-spatial feature extraction, the SS non-Linear Model addresses two primary challenges in hyperspectral imaging: the high-dimensional nature of spectral data and the computational complexities of both spectral and spatial information processing. This model not only compresses hyperspectral data efficiently, but also enhances classification performance through its unique architecture, setting a new benchmark for hyperspectral analysis.

Comprehensive evaluations on three major hyperspectral datasets: Houston, India, and the University of Pavia demonstrate that the SS non-Linear Model consistently outperforms conventional transformer-based and CNN-based approaches. It achieves high accuracy with reduced computational demand, underscoring its robustness and adaptability across diverse landscapes and land cover types. This superior classification performance is achieved without the resource intensity typically associated with advanced transformer models, showcasing the model's efficiency and accuracy trade-offs.

A significant feature of the SS non-Linear Model is its streamlined computational efficiency. Through careful design that minimizes memory requirements and computational operations, the model achieves rapid training and testing, making it suitable for deployment in environments with limited computing power. Experimental results indicate that the SS non-Linear Model achieves similar or improved accuracy with substantially lower FLOPs and parameter counts, thereby opening possibilities for large-scale, real-time hyperspectral data analysis in resource-constrained settings.

This efficiency positions the SS non-Linear Model as a practical choice for applications requiring high-speed, accurate hyperspectral analysis, such as precision agriculture, urban monitoring, and mobile-based remote sensing. By achieving a balance between accuracy and computational demand, the SS non-Linear Model represents a critical advancement in making sophisticated hyperspectral imaging technology accessible and operationally feasible across various platforms.

In conclusion, the SS non-Linear Model stands as a pioneering model that combines bidirectional processing with computational efficiency, advancing the field of hyperspectral imaging. Its design exemplifies the potential of state-space models for high-dimensional spectral data analysis, paving the way for future research into their broader application in remote sensing. This model not only advances hyperspectral imaging capabilities but also inspires further investigation into the use of state-space models for scalable, efficient, and accurate remote sensing data analysis. The results of this study support the role of the SS non-Linear Model as a transformative approach, promising continued exploration of its utility and impact across the spectrum of remote sensing technologies.

% ==============
% # REFERENCES #
% ==============

% {\appendices
% \section*{Proof of the First Zonklar Equation}
% \section*{Proof of the Second Zonklar Equation}
% Appendix two text goes here.}

\bibliographystyle{IEEEtran}

\bibliography{IEEEabrv, Reference. bib}

\vfill

\end{document}